# Planning and Navigation of Climbing Robots in Low-Gravity Environments

Steven Morad
Aerospace and Mechanical Engineering Department
University of Arizona
Tucson, Arizona
smorad@email.arizona.edu

Himangshu Kalita
Aerospace and Mechanical Engineering Department
University of Arizona
Tucson, Arizona
hkalita@email.arizona.edu

Jekan Thangavelautham
Aerospace and Mechanical Engineering Department
University of Arizona
Tucson, Arizona
jekan@email.arizona.edu

*Abstract*— **Advances in planetary robotics have led to wheeled robots that have beamed back invaluable science data from the surface of the Moon and Mars. However, these large wheeled robots are unable to access rugged environments such as cliffs, canyons and crater walls that contain exposed rock-faces and are geological time-capsules into the early Moon and Mars. We have proposed the SphereX robot with a mass of 3 kg, 30 cm diameter that can hop, roll and fly short distances. A single robot may slip and fall, however, a multirobot system can work cooperatively by being interlinked using spring-tethers and work much like a team of mountaineers to systematically climb a slope. We consider a team of four or more robots that are interlinked with tethers in an "x" configuration. Each robot secures itself to a slope using spiny gripping actuators, and one by one each robot moves upwards by crawling, rolling or hopping up the slope. Apart from climbing, path planning, and navigation is another critical challenge that needs to be solved to make the whole approach feasible. For climbing navigation, a multirobot system needs to have up to date info of its location, together with a macroscopic map of the climbing surface and a detailed map ahead. This system with limited sensor range needs to discern and identify feasible pathways to make the next climbing step much like a human mountaineer. These climbing pathways consist of a series of anchor points for the robot to grip onto next. Identifying one or more feasible pathways is a critical challenge as the terrain ahead needs to be acquired, followed by identification and ranking of anchor points to grip. The climbing task resembles a maze with wrong pathways leading to dead-end. The multirobot systems need to autonomously explore climbing pathways and know when to give up. In this paper, we present a human devised autonomous climbing algorithm and evaluate it using a high-fidelity dynamics simulator. The climbing surfaces contain impassable obstacles and some loosely held rocks that can dislodge. Under these conditions, the robots need to autonomously map, plan and navigate up or down these steep environments. Autonomous mapping and navigation capability is evaluated using simulated lasers, vision sensors. The human devised planning algorithm uses a new algorithm called bounded-leg A\*. Our early simulation results show much promise in these techniques and our future plans include demonstration on real robots in a controlled laboratory environment and outdoors in the canyons of Arizona.**

## I. INTRODUCTION

Large areas of the Moon, Mars and asteroids are covered in rugged terrain (Fig. 1) inaccessible by current wheeled rovers [1][2]. On Mars, these areas may hold exposed canyons, cliffs and craters that can provide insight into early geologic history, existence water, and past habitability. Our approach to accessing these sites are to use a network of small, low-cost robotic platforms called SphereX [8],[9],[42],[43]. Each SphereX can hop, roll and fly short distances. Moreover, with the addition of a suitable gripping skin, these robots can grasp onto rough terrain and rest on precarious/sloped surfaces.

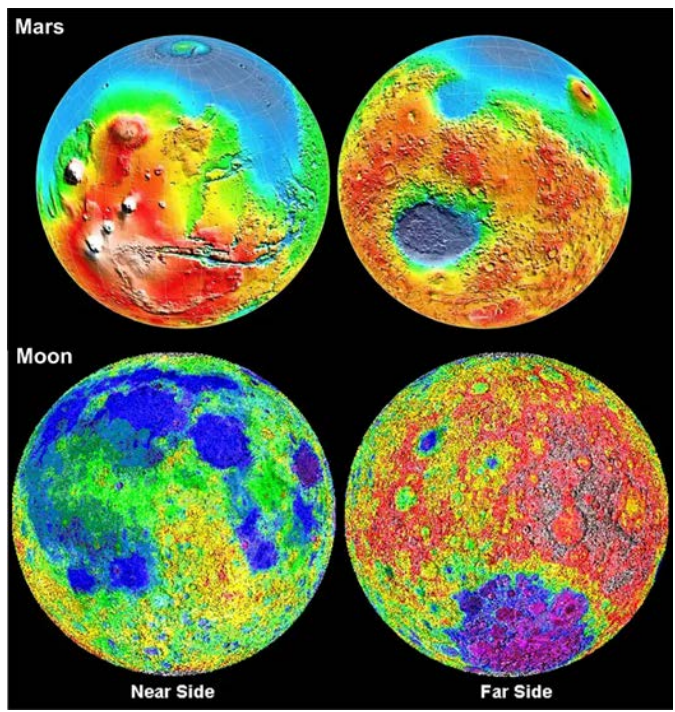

Fig. 1. Topological maps of the Moon and Mars. Note Moon and Mars not shown to scale. Areas shown in orange, red and brown are inaccessible by current rovers.

These robots can climb up a slope by hopping/rolling up a distance and then gripping on the surface. However, a single robot may slip and fall if the gripping mechanism fails to grasp. This can be avoided by developing a multirobot system that can work cooperatively by being interlinked using spring-tethers and work much like a team of alpine mountaineers to systematically climb a slope.

A multirobot team exceeds the sum of its parts by tackling complex slopes that would otherwise be too risky for a single robot to traverse. The multirobot system is comprised of four spherical robots that are interlinked with tethers in an "x" configuration. Each robot is secured to a slope using spiny gripping actuators, and one by one each robot moves upwards by crawling, rolling or hopping up the slope. If any one of the robots loses grip, slips or falls, the remaining robots will be holding it up as they are anchored.

In this paper, we present dynamics and control simulation of an autonomous multirobot system that cooperates to climb sloped surfaces by successively hopping, rolling and crawling. We then extend this approach to incorporate navigation and planning enabling climbing over long distances. This multirobot approach for climbing sloped cliff surfaces holds great potential for exploring cliff and extremely rugged surface environments on Mars, Moon and asteroids. Recent research suggests that wind may have formed channels down the crater slopes in place of water as seen by high-resolution images from the Mars Reconnaissance Orbiter [11]. Getting up-close and traversing these slopes enables going back in time to get a better understanding of the geological history of Mars.

The benefits of the SphereX system can also be realized on milligravity surfaces such as asteroids. There are 150,000+ asteroids, with a large number located in the asteroid belt between Mars and Jupiter [41]. They range in size with diameters ranging from a few tens of meters to several hundred kilometers. On milligravity surfaces, hopping and flying is simple and uses negligible propellant. However, gravity varies throughout the surface and too much thrust can result in a spacecraft achieving escape velocity. Using the proposed multirobot approach with robots anchored to the surface keeps the entire system secure. In the following sections, we present background and related work followed by system overview, dynamic simulations, discussions, conclusions and future work.

## II. RELATED WORK

Numerous robotic systems have been proposed for extreme-environment mobility and exploration but climbing sloped natural terrains has always been a major challenge. Much work has been done in this field to develop tethered, legged and wheeled robotic systems. For exploring volcanoes, Dante II was developed which is an eight-legged walking rover with a tethered rappelling mobility system [24]. Teamed Robots for Exploration and Science on Steep Areas (TRESSA) is a dual-tethered robotic system used for climbing steep cliff faces with slopes varying from 50 to 90 degrees [25]. NASA JPL successfully demonstrated accessing 90-degree vertical cliffs and collecting samples using the Axel platform which is a two-wheeled rover tethered to its host platform [26]. Another example is the All-Terrain Hex-Limbed Extra-Terrestrial Explorer (ATHLETE) rover developed by NASA JPL which has 6-DOF limbs, each attached with a 1-DOF wheel [27].

Several other robotic systems have been developed that uses friction, suction cups, magnets and sticky adhesives to climb sloped terrains. The Legged Excursion Mechanical Utility Rover (LEMUR IIb) developed by NASA JPL is a four-limbed robot capable of free-climbing vertical rocky surfaces, urban rubble piles, sandy terrains and roads using only friction at contact points [28]. Stickybot developed at Stanford employs several design principles adapted from the gecko lizard like hierarchical compliance, directional adhesion and force control to climb smooth surfaces at very low speeds [29]. Spinybot II can climb a wide variety of hard, outdoor surfaces including concrete, stucco, brick and sandstone by employing arrays of microspines that catch on surface irregularities [30]. The Robots in Scansorial Environments (RiSE) is a new class of vertical climbing robots that can climb a variety of human-made and natural surfaces employing a combination of biologically inspired attachments, dynamic adhesion and microspines [31]. NASA JPL has also developed an anchoring foot mechanism for sampling on the surface of near Earth asteroids using microspines that can withstand forces greater than 100 N on natural rock and has proposed to use it on the Asteroid Retrieval Mission (ARM) [32].

MINERVA, the only hopping robot to have actually flown to a low-gravity body had two rudimentary navigation modes. In mode one, MINERVA would hop in the direction of the goal until it reached the goal. In mode two, it would hop away from the sun to prevent overheating [6]. MINERVA II, slated for a launch within the next few years has a very similar a navigation system to MINERVA I [10]. SLIM is another mission proposed by Japan's Aerospace Exploration Agency (JAXA). It will unload two 1.5 kg landers that will drive up to the mouth of a known lava-tube on the moon and look down into the lava tube [12].

Thayer et al. investigate dynamic mapping of an environment using individual robots, applied to urban combat scenarios [13]. The robots proposed are similar the SphereX, and Thayer highlights the problem of obstacles obstructing the view of smaller robots. *Gingras et al.* [14] describes an image processing process very similar to our path planning and navigation algorithm described here. In their process, 2D Delaunay and ball-pivoting methods are used for surface reconstruction [13]. Both methods were attempted in our simulations, but the approach detailed in our path planning section ultimately generated better results. Wu et al. provides a distributed wall climbing system [35]. This robot consists of a mother robot, a controller, and child robots. The mother requires continuous navigation input from a separate host computer and does not do any navigation itself. Bretl et al. discusses a long-term planning algorithm for a multi-limbed free-climbing robot. Using friction and robot-hands, it was able to climb a rock wall using JPL's LEMUR robot as a platform [36]. As discussed with the LEMUR 3 paper, coordinating a climbing robot with multiple arms, each with many degrees of freedom is extremely complex. The algorithm generates an enormous search space for possible poses, gaits, and by extension, paths. While it was able to climb rock walls, this was only possible with human generated graphs of designated anchor points for its robot-hands. Our past work has also shown feasibility of multiple small robots working together as a team [37],[38],[39],[40].

For the proposed multirobot system, the motivation is taken from proven methods used by alpinists to climb mountains. These mountaineers use ice axes and crampons to grip on the surface and climb steep mountain slopes. The use of legs and hands provide four contact points to the sloped surface. The systematic climbing approach is redundant in nature as even when each attempt to grip onto a higher location fails, the climber is still secure with his two feet and one hand gripping tightly onto the slope. Inspired by mountaineers, our approach utilizes a multirobot climbing and flying system that has inherent redundancies to recover from individual slips and falls. Moreover, we provide a way to autonomously search for and designate anchor points.

## III. SYSTEM OVERVIEW

The multirobot system designed to move in vertical/inclined natural sloped consists of four identical spherical robots interlinked together with spring tethers in an "x" configuration. Each robot has a mass of 3 kg and diameter 30 cm. Fig. 2 shows the internal and external views of each SphereX robot. The lower half of the sphere contains the power and propulsion system, with storage tanks for fuel and oxidizer connected to the main thruster. It also contains a 3-axis reaction wheel system for maintaining roll, pitch and yaw angles and angular velocities along x, y and z axes. The propulsion unit provides thrust along +z axis and the reaction wheel system control the attitude and angular velocity of the robot that enables it to perform ballistic hop. Next is the Lithium Thionyl Chloride batteries with specific energy of 500 Wh/kg arranged in a circle. An alternative to batteries are PEM fuel cells. PEM fuel cells are especially compelling as techniques have been developed to achieve high specific energy using solid-state fuel storage systems that promise 2,000 Wh/kg [33],[34]. However, PEM fuel cells require development for a field system in contrast to lithium thionyl chloride that has already been demonstrated on Mars.

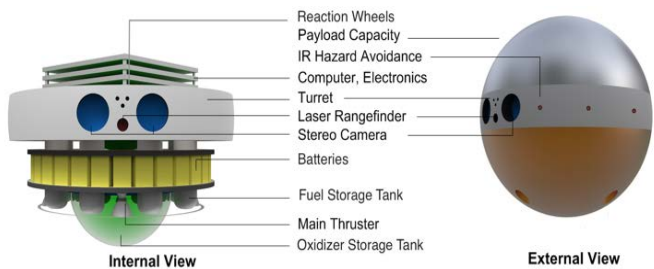

Fig. 2. Internal and external view of each SphereX robot.

For sensing, planning and control, a pair of stereo cameras and a laser range finder are mounted to each robot and they rotate on a turret. This enables the robot to take panoramic pictures and scan the environment without having to move using the propulsion system. Above the turret are two computer boards, IMU and IO-expansion boards, in addition to a power board. The volume above the electronics is reserved for a payload of up to 1 kg. Apart from the proposed propulsion subsystem, all the other hardware components can be readily assembled using Commercial off-the-self (COTS) components.

In the 4-robot climbing system, connections between each robot and tether are made with ball and socket joints. The spring tether introduces three translational degrees of freedom in the system which allows each robot to translate with respect to the others. The ball-socket joint introduces three rotational degrees of freedom in the robot-tether and tether-tether connection, which allows each robot to hop with respect to others. When one robot hops, the other 3 are still engaged to the surface, hence if the hopping robot fails to grip, slips or falls, the other 3 robots will be holding it up as they are anchored. The failed robot will continue to hop until it is able to grip onto the surface at some distance from its initial position. Similarly, the other robots will hop and grip one by one until they attach to a new location.

## IV. CLIMBING SLOPED SURFACES

For climbing sloped surfaces, each spherical robot is equipped with an array of microspines. The robot hops using the propulsion system and reaction wheels and then grips on the rough surface using the array of microspines. However, climbing sloped or vertical cliffs for a single robot is a risky matter. A single robot may slip and fall if the gripping mechanism fails to grasp into the rough surface. However, a multirobot system can work cooperatively by being interlinked using spring-tethers and work much like a mountaineer to systematically climb a slope. We have considered a system of four spherical robots that are interlinked with four spring-tethers in an "x" configuration which work cooperatively to climb a slopped rough surface as shown in Fig. 4.

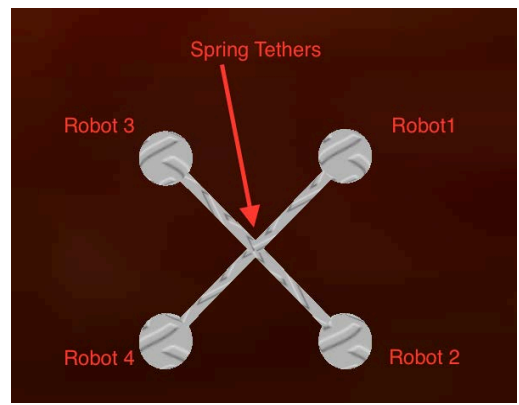

Fig. 3. Cliff climbing multirobot system.

The connections between the robots and tether are made with ball and socket joints. The spring tether introduces one translational degree of freedom in the system which allows each robot to translate with respect to the other robots. The ball-socket joint introduces three rotational degrees of freedom in the robot-tether connection and the tether-tether connection, which allows each robot to hop with respect to the other robots resulting in 3-dimensional movement of the whole system. Fig. 10 shows a Matlab 3D VRML dynamics simulation of a team of 4 robots climbing a slopped surface. In Fig. 10.1 all robots are gripping onto the slope surface. Robot 1 disengages its grip and hops a distance *d* forward and then grips again on the surface. When robot 1 hops, the other three robots are still gripping to the surface, therefore if robot 1 loses grip, slips or falls, the remaining robots will be holding it up as they are anchored. Robot 1 continues to hop until it is able grip onto the surface at a distance *d* from its initial position. Similarly, in Fig. 4.2-10.5 robot 2, 3 and 4 hops and grips on the surface as shown until each robot is displaced by a distance *d*. Fig. 4.5 shows the final configuration of the robot system after it had climbed a distance *d* up the slope.

The surface of each SphereX robot consists of hundreds of microspines. For the robot to climb a wide variety of rough surfaces, it has a combination of large spines as well as smaller spines spread uniformly. Each spine has a shaft diameter of 200-300 μm and a tip radius of 12-25 μm. The maximum load that each spine can sustain per asperity is 1-2 N. Each robot has a mass of 3 kg and each tether has a mass of 0.15 kg, making the mass of the whole system approximately 12.6 kg. On Mars, with a *g* of 3.7 m/s$^2$, the spines need to sustain a load of 47 N. With each spine/asperity contact capable of sustaining 1-2 N load, a minimum of 28 spines should be engaged. With each robot rolling or hopping at a time, the other three robots must share the total load, hence a minimum of 10 spines need be engaged for each robot.

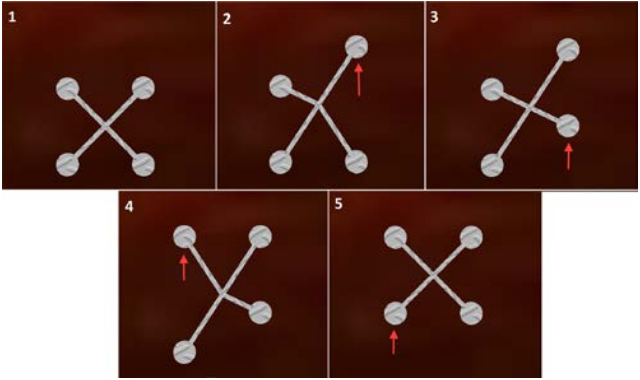

Fig. 4. Sequence of robot movement to climb a steep slope. Each robot hops up the slope, individually and in sequence and grips to the surface. The robots are all attached using spring tether.

Figure 11 shows how the position of each robot and the central tether hub change with time. The initial position of robot 4 is at the origin (0,0,0) and that of robot 1, robot 2 and robot 3 are (1,1,0), (1,0,0) and (0,1,0) (m) respectively. Each robot hops one at a time resulting in the change in position of the instantaneous center. Fig. 12 shows the change in *x*, *y* and *z* coordinates of the "instantaneous center" of the system as its climbing. After four successive hops total, one by each robot, the instantaneous center moves a distance of 0.75 m along *y*-axis in 10 seconds.

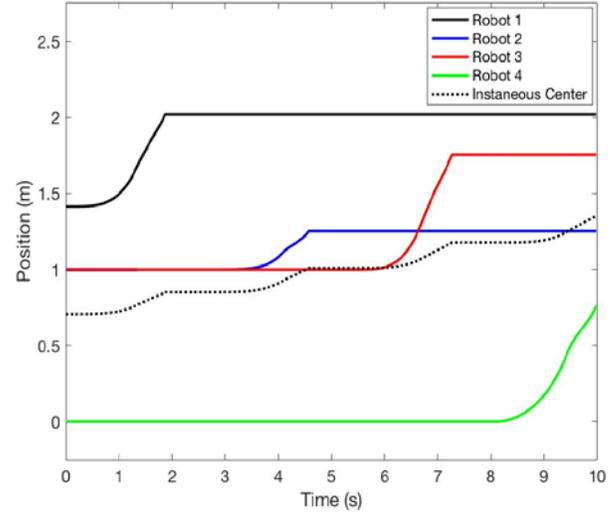

Fig. 5. Change in position of each robot and the instantaneous center during a climb.

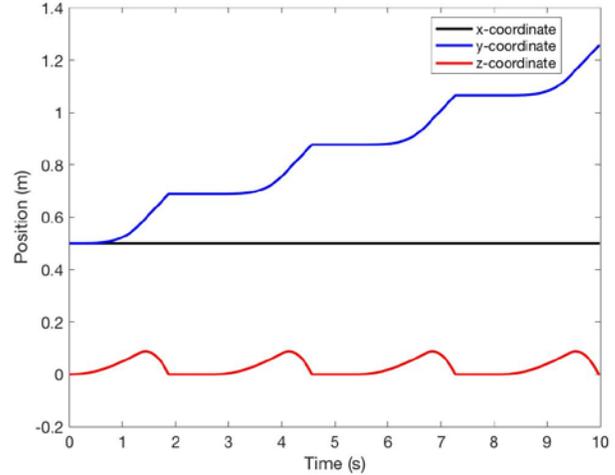

Fig. 6. Change in x, y and z coordinate of the instantaneous center during a climb.

## V. PLANNING AND NAVIGATION FOR CLIMBING

In this section, we extend the climbing techniques shown in the previous section towards exploring an entire slope or a hill. Due to the imaging resolution of current orbiters [11] path planning cannot be done ahead of time and must be done in-situ. Onboard image processing algorithms need to function with incomplete data. Given the height of the SphereX, the terrain can easily obscure the robot's vision (Fig. 8).

Due to small size and power limitations of the SphereX robots, our intent has been to develop a path planning algorithm that is robust and scalable yet can operate on a relatively low-performance microcontroller. We split the navigation problem into two distinct distance-based scales, the local and global scale. The local scale encompasses what is immediately visible to the robot. The global scale will keep track of where the robot has been, and where the robot is in relation to its final target. We will refer to maps of the local area as "scenes", to discern them from the global "map".

We have come up with a real-time process to generate scenes from LIDAR data. This process will also generate a hopping path through the scene in the direction of the target. The process will run individually on each SphereX node, before each hop. The algorithm for the global phase will ultimately run on the scene data provided by this process. A visual overview of the process is presented in Fig. 7. We describe the various phases below.

*A. Scene Mapping and Routing Process*

*1) Point Cloud Generation:* A servo-mounted laser rangefinder is used to generate point cloud data. The sensor uses the round-trip time of the laser to calculate its surroundings. Timing and sensor angle data are converted to points in 3D Cartesian space. Since the SphereX robots are low to the ground, much of the environment will be obscured (Fig. 8). The process is built around the fact that SphereX's vision will be obscured by boulders, hills, and dips. A large part of the process's purpose is to extract as much information as possible from the point cloud.

*2) Normal Recovery:* The following surface reconstruction phase requires normal vectors for each vertex. Unfortunately, the LiDAR scan only provides the coordinates for points, not their orientation. A plane is fit to each vertex and its $k$ neighboring vertices in the point cloud. The normal vector of this plane is then used as the normal vector for the vertex.

*3) Surface Reconstruction:* Due to visual obstructions, the point cloud data from the scanner does not form a continuous mesh. Working with a discontinuous mesh is difficult (Fig. 13), so surface reconstruction is used to fill in gaps and generate a continuous mesh. Screened Poisson Surface Reconstruction (SPSR) [15] is a popular and well-tested method for noisy data and works well for our use case. SPSR generates a continuous mesh of the scan using the computed vertex normals. Here, obscured areas will be filled in, and discontinuities will be smoothed. SPSR is based on an indicator function $\chi$, which will mark a vertex $v$ as either in or not in the surface. Given oriented points (vertices with normals), the gradient $\overline{\nabla_\chi}$ is computed. SPSR computes this by solving the Poisson equation [16]. Additional reconstructive algorithms, such as ball pivoting and VCG were attempted, but they did not guarantee a closed surface like SPSR did. In testing, we found that SPSR performed very well, providing data very close to the actual surface.

*4) Anchor Point Selection:* Once a continuous surface is generated, it can be evaluated for potential points to hop and "anchor" to. In the case of the SphereX, we look for surfaces suitable for microspine grippers. Each vertex will receive a positive score based on how 'risky' it is to anchor to, with a score of zero being the safest. The score will change depending on the type of grippers and the features of the operating environment.

*5) Graph Generation:* A directed graph is formed from the best anchor points to allow for fast analysis and path planning. All anchor points vertices are sorted by ascending score. Anchor points scoring above a threshold are culled to improve graph generation performance. The remaining anchor points are used as vertices in the directed graph. These vertices are potential hop targets. Edges are generated from each anchor vertex to their local neighbors. Each edge contains two different weights. The first weight is the Euclidean distance from vertex $a$ to vertex $b$, denoted as $d_{ab}$. The second weight for a given edge from vertex $a$ to $b$ is $L'_{ab}$, defined as:

$$L'_{ab} = L_b + r_b$$

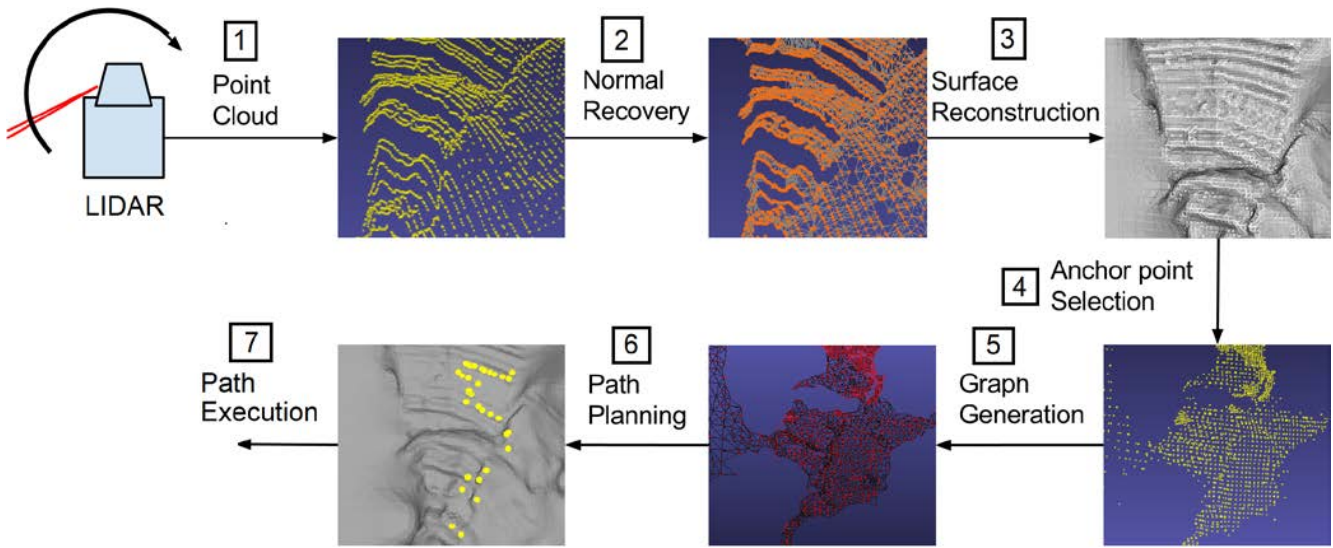

Fig. 7.    Path Planning Algorithm Overview.

where *r* is the score of the vertex computed during anchor point selection. After this phase, the process will be working with the newly created directed graph instead of the surface.

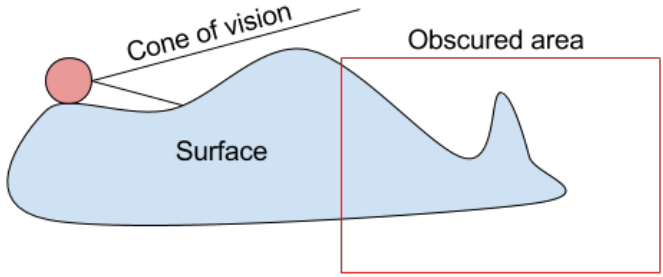

Fig. 8. The SphereX robot shown in red is low to the ground and as a result, areas in front of the robot can be obscured as shown.

*6) Path Planning:* Given a directed graph, path planning in the scene is easy. However, we must remember that our SphereX climbing system consists of four separate robots with a maximum hop distance $h_{max}$ connected to a central point by a limited length spring-tether.

We propose new algorithm based on A*, called bounded-leg A* to keep the four robots with differing potential paths together. A* is a standard algorithm for efficiently finding a path from node *a* to *b* in a graph. Fig. 9 contains a brief explanation of the A* algorithm [17]. Bounded-leg A* ensures that the SphereX nodes never stray too far from each other, and that they only consider hops that are physically possible. With Bounded-leg A*, each node is connected to the central hub with a draw-wire encoder, measuring the current distance from the node to the hub. Using this encoder, bounded-leg A* will provide a series of hops for SphereX to reach the goal without any communication between the nodes.

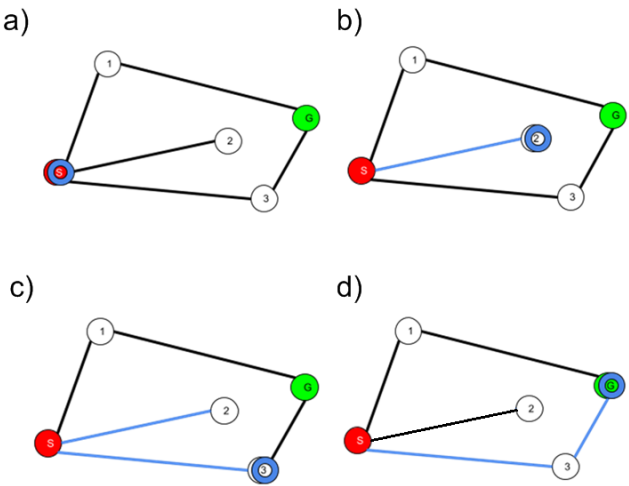

Fig. 9. The A* algorithm generates the shortest path between two vertices. For this scenario, the algorithm moves the blue ring from the red (start) node to the green (goal) node. The edges will turn blue as each is explored. A* picks the path that minimizes the distance between the start and the goal vertices. A* analyzes the path via node 2 (b) but determines it's a dead-end. Then it attempts via node 3 (d) and finds a path to goal. Similarly, it also finds a path via node 1, but find the node 3 path is shorter.
.

Bounded-leg A* is run separately on each SphereX node, generating a path for each SphereX node (Fig. 10 and Fig. 11). Since each iteration is run independently of other nodes, this algorithm can run on a variable number of SphereX nodes. Dealing with a node failure in this case is straightforward. The faulty node can sever its tether, and bounded-leg A* will continue to plan individual paths on the remaining nodes, indifferent to the loss of a node. Bounded-leg A* even makes it possible to detach functional nodes, and have the nodes continue on to separate goals. This can be used to maximize area coverage in a limited amount of time.

```
# Prune edges that
# are too distant
# to hop to
for edge in graph:
    if edge.weight.distance > max_hop_distance:
        graph.discard(edge)
astar(graph)

def astar(graph):
    for edge in vertex.neighbors
        ...
        hub_pos = get_hub_pos()
        if euclidean_distance(
            edge, hub_pos
        ) > max_tether_length:
            graph.discard(edge)
        ...
        cost = edge.weight.loss
        ...
```

Fig. 10. Psuedo-code for the bounded-leg A* based on based on NetworkX's A* [18]

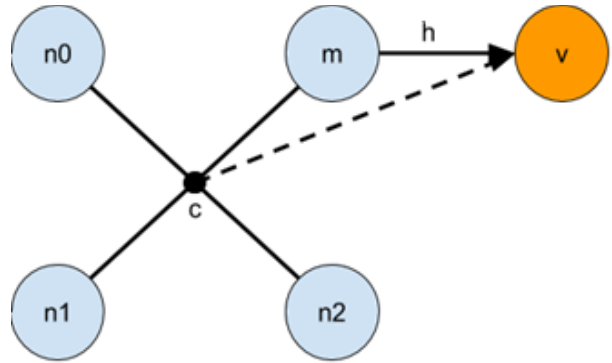

Fig. 11. Four SphereX robots in the x-configuration. Robot *m* hops to anchor point *v* when the node $|cm+h| \leq r_{max}$, where $r_{max}$ is the maximum length of a tether.

*7)  Path Execution:* Each iteration of the above process generates scene data. The scene and onboard inertial measurement data will be fed to a simultaneous localization and mapping (SLAM) algorithm to generate a global map. Given enough scene data, the map will converge to SphereX's surroundings.

D* runs on the global map and provides a path to the goal. D* is an iterative algorithm that will plan a path to the target, and when it discovers an obstruction will replan a new path. D* is the standard for fast robotic path replanning. A visualization of the D* algorithm is presented in Fig. 12. D* will feed directional orders back into the local pipeline. The local and global planner operate on a 2D grid of cells. Each cell represents a scene, and the entire grid represents the global map. In each cell, we have scene data such as rocks, crevices, tunnels, and other environmental features. As we move from cell to cell, we start to fill up the grid with environmental data.

Fig. 12 shows an example of D* in action. The path planning algorithm will naively plan the shortest path to (1, 3), consisting of (1, 1), (1, 2), and (1, 3). The global planner will ask the local planner to move in the southern direction. The local planner scans the cell (1, 1) and plans a series of hops to reach the cell (1, 2). The SphereX executes these hops and replans the path to (1, 2) multiple times within cell (1, 1), but eventually reaches the center of cell (1, 2). As the SphereX robot moves south, it obtains a detailed map of (1, 1) and the map of upper half of (1, 2). The global planner triggers the local planner to move South from (1, 2) to (1, 3). The SphereX starts hopping south towards (1, 3). Unfortunately, after a few hops the local planner notices a giant crevasse between cells (1, 2) and (1, 3) (shown in red) that is too big to jump across. Bounded-leg A* cannot find a solution for (1, 2) to (1, 3). The failure is fed into the global system, which presented as a horizontal "wall" between (1, 2) and (1, 3) to D*. D*, will then re-route around the obstacle, changing the route from (1, 1), (1, 2), (1, 3) to (1, 1), (1, 2), (2, 2), (2, 3), (1, 3).

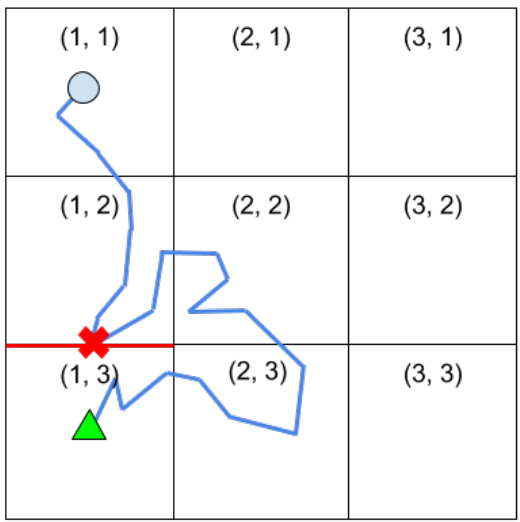

Fig. 12.  SphereX robot using the D* planning algorithm to circumvent an obstruction shown in red.

## VI.  SIMULATION STUDIES

A rock wall was scanned with a LIDAR system with a range of 5.6 m and 3 cm of error, providing a point cloud of over thirty thousand vertices (Fig. 13 left). The scan location was 0.25 m away from the vertical wall, simulating the vertical height of the SphereX. This generated two orders of magnitude more resolution than current orbiters can provide [11],[19]. The 3D point cloud was stretched to simulate a surface that was 500 m tall. Point cloud vertex normals were computed by building the planes presented in Fig. 13 (right). SPSR was applied to point cloud, producing Fig. 14.

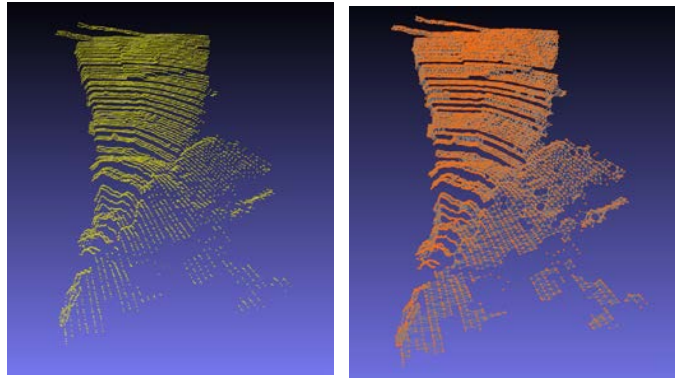

Fig. 13.  (Left) Point cloud of a 5.6 m rock wall composed of over 30,000 vertices. (Right) Microplanes generated to compute vertex normal.

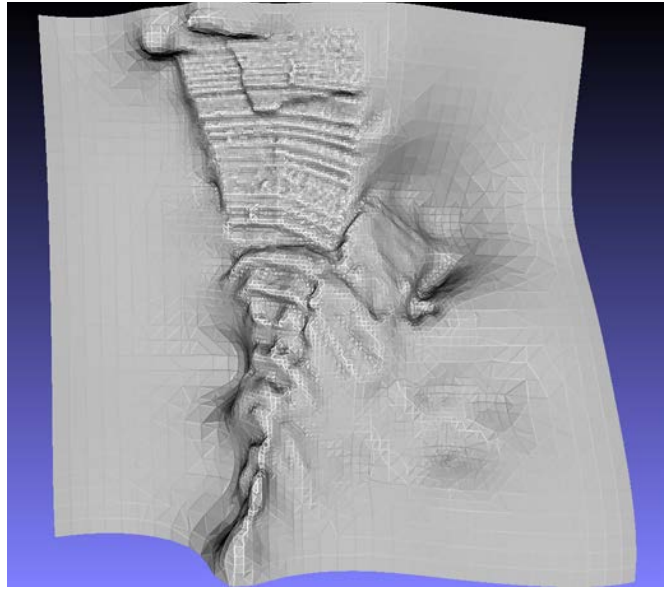

Fig. 14.  3D Surface generated using SPSR.

Anchor points were selected assuming an asteroid surface environment. Two separate criteria were chosen to select anchor points, relative height and surface flatness.

Microspine grippers perform exceptionally poorly on regolith and other sand-like materials [20]. To minimize risk of slipping or sinking, the SphereX nodes must avoid areas with regolith. Generally, depressions such as pits, valleys, and craters will contain deep regolith deposits, as gravity causes regolith to migrate downwards over time [21]. To avoid these areas, we used relative height. This has the added bonus of providing a better vantage point for obtaining better scan data.

We cannot expect to have an accurate gravitational model of non-spherical bodies before visiting and measuring their gravitational fields [22]. While this paper does not focus on dynamics, it's important to note that uncertainty in the gravitational field of a body will lead to uncertainty in the hop trajectory calculations. To help mitigate this risk, we computed the flatness of the surface around an anchor point as a second criterion. If a robot were to hop and miss its target, the hope is that it would not fall into a hole or collide with a jagged rock edge. The best case would be to land in a flat area, clear of obstacles.

We plan to test optical granulometry as a third criterion to assess the regolith conditions at our hop target [23] once our real-world tests begin. The anchor points provided above were used to build the directed graph. The bottom eighty-five percent of vertices were dropped to improve performance, with the remaining fifteen percent forming the graph. Bounded-leg A* was run independently on each SphereX node, producing four separate paths, one for each SphereX node (Fig. 15). The planned route was reviewed by the team and found to be feasible. The planner was also run on Itokawa, the asteroid visited by Hayabusa I and is shown in Fig. 16 [12].

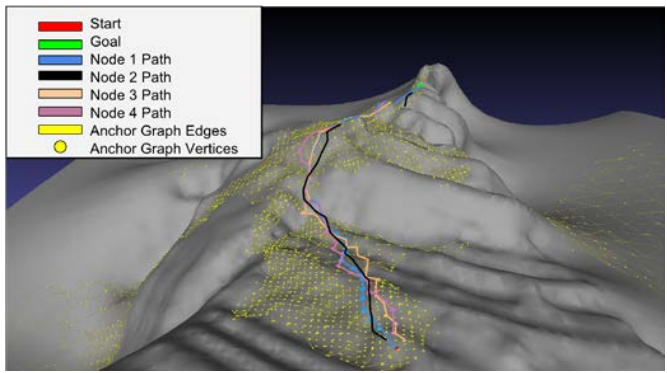

Fig. 15. A path generated by a team of SphereX robots at starting location. The individual paths of each SphereX robot are colored blue, black, brown, and purple respectively. Yellow dots and lines are the vertices and edges respectively in the directed graph generated by the planner. The maximum hop distance per node was 10 m. Note that the path goes behind the mountain into unknown terrain. SPSR has estimated what is behind the mountain, but this may change as the robots get closer and get more accurate data.

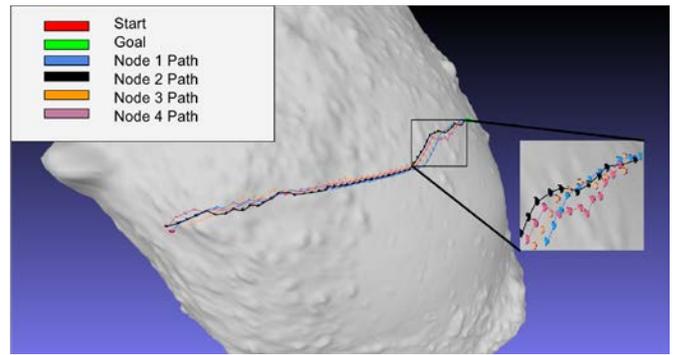

Fig. 16. Path shown for a team of four SphereX robots between two selected locations on asteroid Itokawa. A maximum per-robot hop distance was set to 30 m.

The planner, written mostly in Python, completed in under five seconds on an AMD Ryzen 5 1600. The majority of computational time was spent on anchor point identification. With additional optimization, there is a pathway to execute the planner at 1 Hz, allowing for multiple mid-air vantage points that will greatly improve map coverage. This also opens to door to mid-hop trajectory changes, if desired.

## VII. Conclusions

In this paper we presented a multirobot system for cooperative cliff climbing and steeped surface exploration. We presented a navigation and planning system to enable fully autonomous trekking hundreds of meters at a time. The technique uses a form of the A* algorithm called bounded-leg A* introduced in this paper. The path planning, and navigation algorithm is generic enough that it can be implemented on other hopping robots. In addition, it can handle incomplete and limited 3D data. In this paper, the planner was applied on a 3D topology generated using real LIDAR scans. The results show the algorithm effectively prunes infeasible pathways and selects one of several feasible pathways found. The planner was also applied on 3D models of asteroid Itokawa. Beyond simulation, our efforts are now focused on implementing the algorithm on real robots and testing them in a controlled laboratory environment.

## References

[1] T. Kobayashi, Y. Fujiwara, J. Yamakawa, N. Yasufuku, and K. Omine, "Mobility performance of a rigid wheel in low gravity environments," *Journal of Terramechanics*, vol. 47, no. 4, pp. 261–274, 2010.

[2] R. Siegwart, P. Lamon, T. Estier, M. Lauria, and R. Piguet, "Innovative design for wheeled locomotion in rough terrain," *Robotics and Autonomous systems*, vol. 40, no. 2, pp. 151–162, 2002.

[3] J. Saito, H. Miyamoto, R. Nakamura, M. Ishiguro, T. Michikami, A. M. Nakamura, H. Demura, S. Sasaki, N. Hirata, C. Honda, A. Yamamoto, Y. Yokota, T. Fuse, F. Yoshida, D. J. Tholen, R. W. Gaskell, T. Hashimoto, T. Kubota, Y. Higuchi, T. Nakamura, P. Smith, K. Hiraoka, T. Honda, S. Kobayashi, M. Furuya, N. Matsumoto, E. Nemoto, A. Yukishita, K. Kitazato, B. Dermawan, A. Sogame, J. Terazono, C. Shinohara, and H. Akiyama, "Detailed images of asteroid 25143 itokawa from hayabusa," *Science*, vol. 312, no. 5778, pp. 1341–1344, 2006.

[4] G. Genta, *Mobility on Mars*. Cham: Springer International Publishing, 2017, pp. 218–238.

[5] M. Pavone, J. C. Castillo-Rogez, I. A. D. Nesnas, J. A. Hoffman, N. J. Strange, "Spacecraft/Rover Hybrids for the Exploration of Small Solar System Bodies" *IEEE Aerospace Conference*, 2013.

[6] T. Yoshimitsu, T. Kubota, I. Nakatani, T. Adachi, and H. Saito, "Microhopping robot for asteroid exploration," *Acta Astronautica*, vol. 52, no. 2, pp. 441–446, 2003.

[7] T. Kubota and T. Yoshimitsu, "Intelligent rover with hopping mechanism for asteroid exploration," in *2013 6th International Conference on Recent Advances in Space Technologies (RAST)*, June 2013, pp. 979–984.

[8] H. Kalita and J. Thangavelautham, "Multirobot cliff climbing on low gravity environments," *11th NASA/ESA Conference on Adaptive Hardware and Systems, Pasadena, CA*, 2017.

[9] H. Kalita, R. T. Nallapu, A. Warren, and J. Thangavelautham, "Guidance, navigation and control of multirobot systems in cooperative cliff climbing," *Advances in Astronautical Sciences*, February 2017.

[10] T. Yoshimitsu, T. Kubota, A. Tomiki, and Y. Kuroda, "Development of hopping rovers for a new challenging asteroid," in *Proceedings of the 12th International Symposium on Artificial Intelligence, Robotics and Automation in Space*, 2014, pp. 5C–01.

[11] A. S. McEwen, E. M. Eliason, J. W. Bergstrom, N. T. Bridges, C. J. Hansen, W. A. Delamere, J. A. Grant, V. C. Gulick, K. E. Herkenhoff, L. Keszthelyi *et al.*, "Mars reconnaissance orbiter's high resolution imaging science experiment (hirise)," *Journal of Geophysical Research: Planets*, vol. 112, no. E5, 2007.

[12] J. Haruyama, S. Sawai, T. Mizuno, T. Yoshimitsu, S. Fukuda, and I. Nakatani, "Exploration of lunar holes, possible skylights of underlying lava tubes, by smart lander for investigating moon (slim)," *Transactions of The Japan Society for Aeronautical and Space Sciences, Aerospace Technology Japan*, vol. 10, no. ists28, pp. Pk 7–Pk 10, 2012.

[13] S. M. Thayer, M. B. Dias, B. Nabbe, B. L. Digney, M. Hebert, and A. Stentz, "Distributed robotic mapping of extreme environments," in *Mobile Robots XV and Telemanipulator and Telepresence Technologies VII*, vol. 4195. International Society for Optics and Photonics, 2001, pp. 84–96.

[14] D. Gingras, T. Lamarche, J. L. Bedwani, and . Dupuis, "Rough terrain reconstruction for rover motion planning," in *2010 Canadian Conference on Computer and Robot Vision*, May 2010, pp. 191–198.

[15] M. Kazhdan and H. Hoppe, "Screened poisson surface reconstruction," *ACM Transactions on Graphics (TOG)*, vol. 32, no. 3, p. 29, 2013.

[16] M. Kazhdan, M. Bolitho, and H. Hoppe, "Poisson surface reconstruction," in *Eurographics Symposium on Geometry Processing*, 2006.

[17] A. Hagberg, P. Swart, and D. S Chult, "Exploring network structure, dynamics, and function using networkx," Los Alamos National Laboratory (LANL), Tech. Report, 2008.

[18] A. A. Hagberg, D. A. Schult, and P. J. Swart, "Exploring network structure, dynamics, and function using NetworkX," in *Proceedings of the 7th Python in Science Conference (SciPy2008)*, Pasadena, CA USA, Aug. 2008, pp. 11–15.

[19] J. Saito, H. Miyamoto, R. Nakamura, M. Ishiguro, T. Michikami, A. Nakamura, H. Demura, S. Sasaki, N. Hirata, C. Honda *et al.*, "Detailed images of asteroid 25143 itokawa from hayabusa," *Science*, vol. 312, no. 5778, pp. 1341–1344, 2006.

[20] K. S. Witkoe, "Material testing for robotic omnidirectional anchor," *NASA JPL Technical Report*, 2012.

[21] H. Miyamoto, H. Yano, D. J. Scheeres, S. Abe, O. Barnouin-Jha, A. F. Cheng, H. Demura, R. W. Gaskell, N. Hirata, M. Ishiguro *et al.*, "Regolith migration and sorting on asteroid itokawa," *Science*, vol. 316, no. 5827, pp. 1011–1014, 2007.

[22] D. Scheeres, R. Gaskell, S. Abe, O. Barnouin-Jha, T. Hashimoto, J. Kawaguchi, T. Kubota, J. Saito, M. Yoshikawa, N. Hirata *et al.*, "The actual dynamical environment about itokawa," in *AIAA/AAS Astrodynamics Specialist Conference and Exhibit, Keystone, Colorado*, 2006, pp. 21–24.

[23] N. H. Maerz, T. C. Palangio, and J. A. Franklin, "Wipfrag image based granulometry system," in Proceedings of the *FRAGBLAST 5 Workshop on Measurement of Blast Fragmentation*, Montreal, Quebec, Canada. AA Balkema, 1996, pp. 91–99.

[24] J. E. Bares, D. S. Wettergreen, "Dante II: Technical descriptions, results, and lessons learned" *International Journal of Robotics Research*, July 1999.

[25] T. Huntsberger, A. Stroupe, H. Aghazarian, M. Garrett, P. Younse, M. Powell, "TRESSA: Teamed Robots for Exploration and Science on Steep Areas," *Journal of Field Robotics*, 2007.

[26] I. A. D. Nesnas, P. Abad-Manterola, J. Edlund, J. Burdick, "Axel Mobility Platform for Steep Terrain Excursion and Sampling on Planetary Surfaces," *IEEE Aerospace Conference*, March 2007.

[27] M. Heverly, J. Mattews, M. Frost, C. McQuin, "Development of the Tri-ATHLETE Lunar vehicle prototype" *40th Aerospace Mechanics Symposium*, NASA Kennedy Space Center, May 2010.

[28] T. Bretl, S. Rock, J. C. Latombe, B. Kennedy, H. Aghazarian, "Free-Climbing with a Multi-Use Robot," *Experimental Robotics IX*. Springer Tracts in Advanced Robotics, vol. 21, Springer, Berlin, Heidelberg.

[29] S. Kim, M. Spenko, S. Trujillo, B. Heyneman, D. Santos, M. R. Cutkosky, "Smooth Vertical Surface Climbing with Directional Adhesion," *IEEE Transactions on Robotics*, vol. 24, no. 1, February 2008.

[30] A. T. Asbeck, S. Kim, M. R. Ctkosky, W. R. Provancher, M. Lanzetta, "Scaling hard vertical surfaces with compliant microspine arrays," *International Transactions on Robotics*, vol. 24, no. 1, February 2008.

[31] A. Saunders, D. I. Goldman, R. J. Full, M. Buehler, "The RiSE Climbing Robot: Body and Leg Design," *Unmanned Systems Technology VIII*, vol. 6230, 2006.

[32] A. Parness, "Anchoring Foot Mechanisms for Sampling and Mobility in Microgravity," *IEEE International Conference on Robotics and Automation*, May 2011.

[33] J. Thangavelautham and S. Dubowsky, "On the Catalytic Degradation in Fuel Cell Power Supplies for Long-Life Mobile Field Sensors," *Journal of Fuel Cells: Fundamental to Systems*, pp. 181-195, 2013.

[34] D. Strawser, J. Thangavelautham, S. Dubowsky, "A passive lithium hydride hydrogen generator for low power fuel cells for long-duration sensor networks," *International Journal of Hydrogen Energy*, 2014.

[35] S. Wu, M. Li, S. Xiao, and Y. Li, "A wireless distributed wall climbing robotic system for reconnaissance purpose," in *Mechatronics and Automation, Proceedings of the 2006 IEEE International Conference on. IEEE*, 2006, pp. 1308–1312.

[36] T. Bretl, "Motion planning of multi-limbed robots subject to equilibrium constraints: The free-climbing robot problem," The *International Journal of Robotics Research*, vol. 25, no. 4, pp. 317–342, 2006.

[37] H. Kalita, E. Asphaug, S. Schwartz, J. Thangavelautham, "Network of Nano-Landers for In-Situ characterization of asteroid impact studies," *68th International Astronautical Congress (IAC)*, Australia, 2017

[38] H. Kalita, A. Ravindran, J. Thangavelautham, "Exploration and Utilization of Asteroids as Interplanetary Communication Relays," *IEEE Aerospace Conference* 2018.

[39] J. Thangavelautham, N. Abu El Samid, P. Grouchy, E. Earon, T. Fu, N. Nagrani, G.M.T. D'Eleuterio, "Evolving Multirobot Excavatio Controllers and Choice of Platforms Using Artificial Neural Tissue Controllers," *IEEE Symposium on Computational Intelligence for Robotics and Automation*, 2009.

[40] J. Thangavelautham, K. Law, T. Fu, N. Abu El Samid, A. Smith, G.M.T. D'Eleuterio, "Autonomous Multirobot Excavation for Lunar Applications," *Robotica*, pp. 1-39, 2017

[41] Ivezic, Zeljko, et al. "Asteroids observed by the Sloan Digital Survey." *Survey and Other Telescope Technologies and Discoveries*. Vol. 4836. International Society for Optics and Photonics, 2002.

[42] J. Thangavelautham, M. S. Robinson, A. Taits, T. J. McKinney, S. Amidan, A.Polak, "Flying, hopping Pit-Bots for cave and lava tube exploration on the Moon and Mars," 2nd International Workshop on Instrumentation for Planetary Missions, NASA Goddard, Greenbelt, Maryland, 2014.

[43] H. Kalita, R. T. Nallapu, A. Warren, J. Thangavelautham, "GNC of the SphereX robot for extreme environmentexploration on Mars," Advances in the Astronautical Sciences, February 2017.